%% file: acl_latex.tex
\pdfoutput=1

\documentclass[11pt]{article}

\usepackage{acl}
\usepackage{times}
\usepackage{latexsym}

\usepackage[T1]{fontenc}

\usepackage[utf8]{inputenc}

\usepackage{microtype}

\usepackage{multirow}
\usepackage{booktabs}
\usepackage{graphicx}
\usepackage{subfigure}
\usepackage{pifont}
\usepackage{xspace}
\usepackage{tabularx}
\usepackage{amsmath}
\usepackage{amssymb}
\usepackage{tablefootnote}
\usepackage{algorithm}
\usepackage[noend]{algpseudocode}
\usepackage{amssymb}
\usepackage{soul}
\usepackage{colortbl}
\usepackage{svg}
\usepackage{hyperref}

\input{math_commands}

\newcommand{\datawiki}[0]{Wikidata5M\xspace}
\newcommand{\datafb}[0]{FB15K-237\xspace}
\newcommand{\datawnrr}[0]{WN18RR\xspace}
\newcommand{\dataicews}[0]{ICEWS14\xspace}
\newcommand{\dataicewss}[0]{ICEWS05-15\xspace}

\newcommand{\method}[0]{CSProm-KG\xspace}

\definecolor{mygray}{gray}{.9}

\title{Dipping PLMs Sauce: Bridging Structure and Text for Effective Knowledge Graph Completion via Conditional Soft Prompting}

\author{Chen Chen$^{1}$, Yufei Wang$^{2}$, Aixin Sun$^{1}$, Bing Li$^{3,4}$ \and Kwok-Yan Lam$^{1}$\thanks{~~Corresponding author} \\
Nanyang Technological University, Singapore$^1$ \\
Macquarie University, Sydney, Australia$^2$ \\
IHPC$^3$ and CFAR$^4$, Agency for Science, Technology and Research (A*STAR), Singapore\\
\texttt{\{S190009,axsun,kwokyan.lam\}@ntu.edu.sg},\\
\texttt{yufei.wang@students.mq.edu.au} \\
\texttt{li\_bing@cfar.a-star.edu.sg} \\
}

\begin{document}
\maketitle
\begin{abstract}
Knowledge Graph Completion (KGC) often requires both KG structural and textual information to be effective. Pre-trained Language Models (PLMs) have been used to learn the textual information, usually under the fine-tune paradigm for the KGC task. However, the fine-tuned PLMs often overwhelmingly focus on the textual information and overlook structural knowledge. To tackle this issue, this paper proposes \method (\textbf{C}onditional \textbf{S}oft \textbf{Prom}pts for \textbf{KG}C) which maintains a balance between structural information and textual knowledge. \method only tunes the parameters of \emph{Conditional Soft Prompts} that are generated by the entities and relations representations. We verify the effectiveness of \method on three popular static KGC benchmarks \datawnrr, \datafb and \datawiki, and two temporal KGC benchmarks \dataicews and \dataicewss. \method outperforms competitive baseline models and sets new state-of-the-art on these benchmarks. We conduct further analysis to show (i) the effectiveness of our proposed components, (ii) the efficiency of \method, and (iii) the flexibility of \method~\footnote{Our source code is available at~\url{https://github.com/chenchens190009/CSProm-KG}}.
\end{abstract}

\input{sections/1.Introduction}

\input{sections/2.Related_work}

\input{sections/3.Method}

\input{sections/5.Result}

\input{sections/7.Conclusion}

\input{sections/9.Limitations}

\input{sections/10.Acknowledgement}

\bibliography{anthology,custom}
\bibliographystyle{acl_natbib}

\clearpage
\appendix
\input{sections/8.Appendix}

\end{document}

%% file: math_commands.tex
\usepackage{amsmath,bm}
\usepackage{amsfonts}

\def\eqref#1{equation~\ref{#1}}

\def\1{\bm{1}}

\def\vh{{\bm{h}}}

\def\vs{{\bm{s}}}

\def\vw{{\bm{w}}}

\DeclareMathAlphabet{\mathsfit}{\encodingdefault}{\sfdefault}{m}{sl}
\SetMathAlphabet{\mathsfit}{bold}{\encodingdefault}{\sfdefault}{bx}{n}

\newcommand{\R}{\mathbb{R}}



%% file: sections/1.Introduction.tex
\section{Introduction}


Knowledge Graphs (KGs) have both complicated graph structures and rich textual information over the facts. Despite being large, many facts are still missing. Knowledge Graph Completion (KGC) is a fundamental task to infer the missing facts from the existing KG information.


\begin{figure}[!t] 
    \centering 
    \includegraphics[width=\columnwidth]{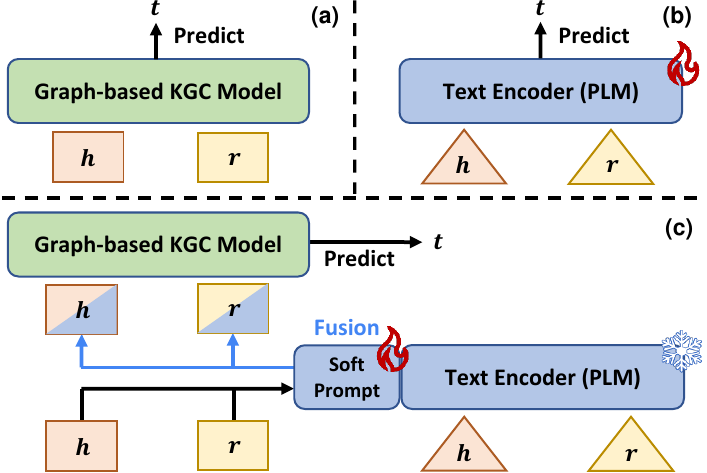} 
    \caption{Given head entity $h$ and relation $r$, KGC is to find out the true tail entity $t$. Graph-based KGC models represent $h$ and $r$ as embeddings (rectangular boxes) to learn the KG structure information (Figure.a).  PLM-based KGC models only feed the textual knowledge (triangle boxes) of $h$ and $r$ into the Pre-trained Language Model (PLM) to predict the missing entity (Figure.b). \method fuses both types of information via the \emph{Soft Prompt} and uses a graph-based KGC model to make the final prediction (Figure.c).} 
    \label{fig:intro figure} 
\end{figure}

Graph-based KGC models~\cite{TransE, DistMult, ConvE} represent entities and relations using trainable embeddings. These models are trained to keep the connections between entities and relations over structural paths, and tail entities are inferred via various transitional relations. Despite being effective in modelling KG structural information, these methods are unable to incorporate linguistic context. Recently, pre-trained language models (PLMs) are applied to fill up this gap~\cite{KG-BERT, StAR, GenKGC}. The proposed solutions often directly fine-tune the PLMs to choose the correct entities either relying on pure textual context or using structural add-ons as a complementary~\cite{StAR}. However, 
PLMs are normally equipped with large-scale parameters and  linguistic inherence obtained from their pre-training stage. As a result, these PLM-based models remain overwhelmingly focusing on the textual information in KGs and tend to overlook the graph structure. 
For example, given an incompleted fact \emph{(Mona Lisa, painted by, ?)},  the PLM-based models may confuse between \emph{Leonardo DiCaprio} and \emph{Leonardo da Vinci} simply because they are textually similar. 
Thus, in this paper, we focus on the research question: \emph{Can we effectively fuse the KG structural information into the PLM-based KGC models?}

To this end, we propose a novel \method model (\textbf{C}onditional \textbf{S}oft \textbf{Prom}pts for \textbf{KG}C) which is a structure-aware frozen PLMs that could effectively complete the KGC task. The core of \method is \emph{Conditional Soft Prompt} that is an structure-aware version of \emph{Soft Prompt}~\cite{prefix-tuning, Prompt-tuning}. Previously, \emph{Soft Prompt} is a sequence of \emph{unconditional} trainable vectors that are prepended to the inputs of frozen PLMs. Such design could effectively avoid the issue of over-fitting towards textual information caused by fine-tuning and allow the frozen PLMs to learn the downstream tasks~\citep{wang-etal-2022-promda}. However, such naive \emph{Soft Prompts} cannot represent any structural information in KG. To remedy this, as shown in Figure~\ref{fig:intro figure} (c), we propose the prompt vectors \emph{conditioned} on the KG entities and relations embeddings. Specifically, we use the entity and relation embeddings to generate \emph{Conditional Soft Prompts} which are then fed into the frozen PLMs to fuse the textual and structural knowledge together. The fused \emph{Conditional Soft Prompts} are used as inputs to the graph-based KGC model that produces the final entity ranking results. We further propose \emph{Local Adversarial Regularization} to improve \method to distinguish textually similar entities in KG.

We evaluate \method on various KGC tasks and conduct experiments on \datawnrr, \datafb and \datawiki for Static KGC (SKGC), and on \dataicews and \dataicewss for Temporal KGC (TKGC). \method outperforms a number of competitive baseline models, including both graph-based and PLM-based models. We conduct ablation studies to show the strength of prompt-based methods against the fine-tuning counterparts and the effectiveness of each proposed components. We also demonstrate the flexibility of \method with different graph-based models, and the training and inference efficiency of \method.

%% file: sections/2.Related_work.tex
\section{Related Work}

\paragraph{Graph-based methods}
Graph-based methods represent each entity and relation with a continuous vector by learning the KG spatial structures. They use these embeddings to calculate the distance between the entities and KG query to determine the correct entities. The training objective is to assign higher scores to true facts than invalid ones. In static KGC (SKGC) task, there are two types of methods: 1) Translational distance methods measure the plausibility of a fact as the distance between the two entities, ~\cite{TransE, TransR, TransH}; 2) Semantic matching methods calculate the latent semantics of entities and relations~\cite{RESCAL, DistMult, ConvE}. In temporal KGC (TKGC) task, the systems are usually based on SKGC methods, with additional module to handle KG factual tuples timestamps~\cite{HyTE, DE-SimplE, TKGCframework}.

\paragraph{PLM-based methods}
PLM-based methods represent entities and relations using their corresponding text. These methods introduce PLM to encode the text and use the PLM output to evaluate the plausibility of the given fact. On SKGC, \citet{KG-BERT} encode the combined texts of a fact, then a binary classifier is employed to determine the plausibility. To reduce the inference cost in~\citet{KG-BERT}, \citet{StAR} exploit Siamese network to encode $(h, r)$ and $t$ separately. Unlike previous encode-only model, \citet{GenKGC, KGT5} explore the \emph{Seq2Seq} PLM models to directly generate target entity text on KGC task.

\paragraph{Prompt tuning}
~\citet{GPT3} first find the usefulness of prompts, which are manually designed textual templates, in the GPT3 model.~\citet{AdvTrigger, AutoPrompt} extend this paradigm and propose hard prompt methods to automatically search for optimal task-specific templates. However, the selection of discrete prompts involves human efforts and difficult to be optimized together with the downstream tasks in an end-to-end manner.~\cite{prefix-tuning, Prompt-tuning}
relax the constraint of the discrete template with trainable continuous vectors (soft prompt) in the frozen PLM. As shown in~\citet{prefix-tuning, Prompt-tuning, P-tuning}, frozen PLM with \emph{Soft Prompt} could achieve comparative performance on various NLP tasks, despite having much less parameters than fully trainable PLM models. To the best of our knowledge, we are the first to apply \emph{Soft Prompt} to PLM-based KGC model.

%% file: sections/3.Method.tex
\begin{figure*}[!ht] 
    \centering 
    \includegraphics[width=\linewidth]{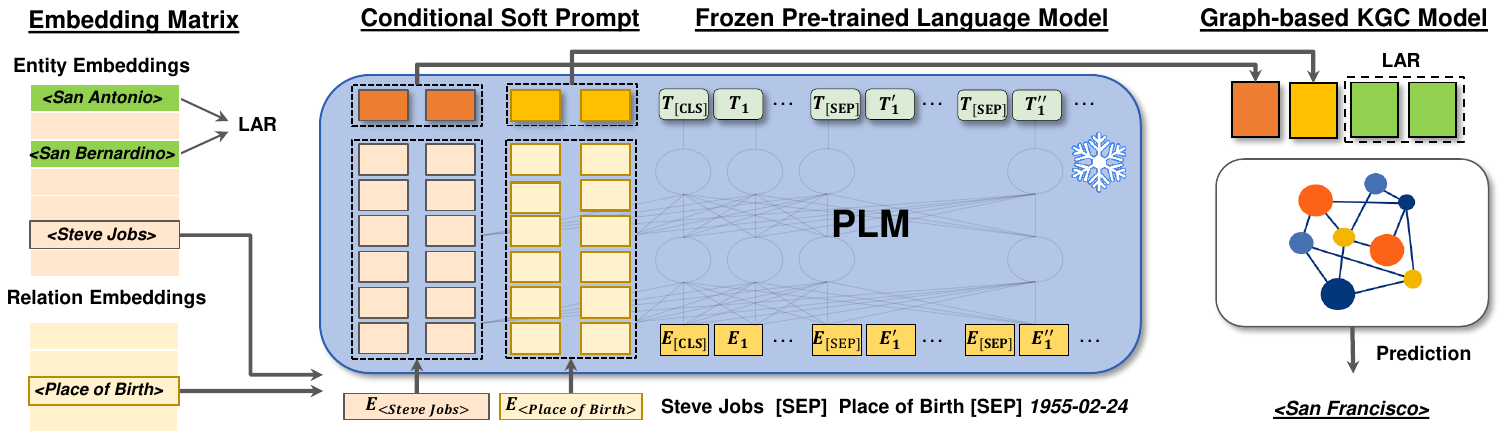} 
    \caption{An example of \method for the KG query (\emph{Steve Jobs}, \emph{Place of Birth}, ?, 1955-02-24). \method uses the embeddings of entities and relations (randomly initialized before training) to generate \emph{Conditional Soft Prompt}. In the frozen PLMs, \emph{Conditional Soft Prompt} fully interacts with the textual information of the KG queries. The outputs are fed into graph-based KGC model to make the final prediction. To improve \method's ability in distinguishing textually similar entities, we further add \emph{LAR} examples that are similar to the tail entities during training.  \method effectively learns both structural and textual knowledge in KG.} 
    \label{fig:model architecture} 
\end{figure*}
\section{Method}
\label{methodsec}
In this section, we first formulate \emph{Knowledge Graph Completion} in Sec.~\ref{sec:setting}. We then introduce \method in Sec.~\ref{methodoverview} to Sec.~\ref{objective}.

\subsection{Knowledge Graph Completion}
\label{sec:setting}
Knowledge graph (KG) is a directed graph with a collection of fact tuples. Let $T = \{V, R, L, M\}$ be a KG instance, where $V$, $R$, $L$ and $M$ denote the entity, relation, edge (fact) and meta information set respectively. Each edge $e \in L$ is $(h, r, t, m) \in V \times R \times V \times M$ which connects head entity  $h$ and target entity $t$ with relation type $r$, and is associated with meta information $m$. In Static KGs (SKG), no meta information is involved (i.e. $M=\emptyset$). In Temporal KGs (TKG), each fact has a corresponding timestamp and $M$ includes all fact timestamps. \emph{Knowledge Graph Completion} (KGC) is to predict the target entity for KG queries $(h, r, ?, m)$. The queries $(?, r, t, m)$ are converted into $(t, r^{-1}, ?, m)$, where $r^{-1}$ is the inverse of $r$. In this paper, \method learns a score function $f(h, r, t, m) : V \times R \times V \times M \rightarrow V$ that assigns a higher score for valid facts than the invalid ones. 


\subsection{\method Overview}
\label{methodoverview}
Motivated by the observation that \emph{Soft Prompts} in a frozen PLM is effective in solving the over-fitting issue~\cite{wang-etal-2022-promda}, we apply \emph{Soft Prompts} in \method to avoid the KGC models overly focusing on the textual information. 
Although several research initiatives have explored the utilization of both structural and textual information for NLP tasks~\cite{10.1145/3511808.3557452,xiao-etal-2021-bert4gcn}, none of them is capable of solving the over-fitting issue over textual information in the context of KGC.
Figure~\ref{fig:model architecture} shows the architecture of \method which includes three important components: a fully trainable \emph{Graph-based KGC model} $G$, a frozen Pre-trained language model (PLM) $P$, and a trainable \emph{Conditional Soft Prompt} $S$. Firstly, the embeddings in $G$, which are \emph{explicitly} trained to predict entities using structural knowledge, are used to generate the parameters of $S$. In this way, $S$ is equipped with KG structural knowledge. We then feed the generated $S$, as well as the corresponding text of entities and relations, into $P$. Finally, the PLM outputs of $S$ are extracted as the final inputs to $G$ which produces final results for the KGC tasks. This allows the structural knowledge from $G$ and the textual knowledge from $P$ to be equally fused via $S$. To further improve the robustness of \method, we propose \emph{Local Adversarial Regularization}, which selects textually similar entities for training to be detailed shortly. 



\subsection{Graph-based KGC Model $G$}
\label{kgcmodel}
In \method, the graph-based KGC models $G$ represents KG entities and relations as continuous embeddings. 
Given a KG query $(h, r, ?, m)$, we represent $h$ and $r$ as embeddings $E_{e}$ and $E_{r} \in \R^{d}$ where $d$ is the embedding size. $E_{e}$ and $E_{r}$ are used at both \emph{inputs} and \emph{outputs}. At \emph{inputs}, we use these embeddings to generate \emph{Conditional Soft Prompt} which further interacts with the textual inputs of the frozen PLM $P$. At \emph{outputs}, we use these embeddings to calculate $f(h, r, t, m)$ which produces the entity ranking for KG queries. For example, when using ConvE as $G$, the corresponding $f(h, r, t, m)$ is the dot-product between the representation of $(h, r)$ and the tail entity embeddings. Note that, \method is flexible enough to work well with any existing graph-based KGC models. We will show this flexibility in Sec.~\ref{sec:flexibility}.

\subsection{Pre-trained Language Model $P$}
Let's assume that the pre-trained language model $P$ has $l$ transformer layers with hidden size $H$. To represent a KG query $(h, r, ?, m)$, we jointly represent $h$, $r$ and $m$ by extracting and concatenating their corresponding raw tokens, including their names and their corresponding descriptions if available. We connect the texts of $h$ and $r$ with a special token [SEP], and feed the joint text into the frozen PLM $P$. For TKGC tasks, we simply add the event timestamp after the joint text of $h$ and $r$. We show the effectiveness of this design choice in Sec.~\ref{sec:ablation}.

\subsection{Conditional Soft Prompt $S$}
\label{sec:prompt tuning}


\emph{Soft Prompt} prepends a sequence of trainable embeddings at the inputs to a frozen Pre-trained Language model. ~\citet{prefix-tuning} propose \emph{Layer-wise Soft Prompt} which inserts relatively short prompt sequences (e.g., 5 - 10 vectors) at each layer and allows frequent interaction with the entities' and relations' textual information in PLMs. Inspired by this, we propose a novel \emph{Conditional Soft Prompt} which has $k$ trainable vectors on each layer. Specifically, the $i^{th}$ input for the $j^{th}$ layer $\vh^j_i \in \R^{H}$ is defined as:
\begin{equation}
\vh^j_i=\left\{
\begin{array}{ccl}
\vs^j_i & & {i \leq k}  \\
\vw_i & & {(i > k) \wedge (j = 0)} \\
\mathit{Trans}(\vh^{j-1}_{:})_i & & {\text{Otherwise}}
\end{array} \right.
\end{equation}
where $\mathit{Trans}(\cdot)$ is the forward function of Transformer layer in $P$, $w_i$ is the fixed input word embedding vector and $\vs^j_i$ is the $i^{th}$ prompt vector at $j^{th}$ layer. The $\mathit{Trans}(\cdot)$ works on the entire sequence (prompt + text). \emph{Conditional Soft Prompt} is designed to connect with embeddings in $G$, we use the embeddings of entities and relations $E_{e}$ and $E_{r}$ to generate \emph{Conditional Soft Prompt} $S$. Formally,
\begin{align}
S &= [F(E_{e}); F(E_{r})]  \\
F(x) &= W_{out} \cdot (\text{ReLU}(W_{in} \cdot x))
\end{align}
where $W_{in} \in \R^{d_h \times d}$ and $W_{out} \in \R^{(l * H * k/2) \times d_h}$ are trainable weight matrices and $d_h$ is the middle hidden size for the mapping layers. We then re-organize $F(E_{e})$ and $F(E_{r})$ into a sequence of input embeddings and evenly distribute them into each PLM layer. In this process, the input tokens for $P$ and \emph{Conditional Soft Prompt} $S$ are fully interacted with each other, allowing the structural knowledge in $G$ (linearly mapped to $S$) and textual knowledge in $P$ to be fully fused together. 

\subsection{Local Adversarial Regularization} 
\label{sec:LAR}
As PLMs are frozen, the model may lose part of flexibility in distinguishing textually similar entities via tuning of the Transformer layers. To enhance \method's ability to distinguish textually similar entities, inspired by~\cite{FGSM}, we introduce an Adversarial Regularization term. Different from conventional adversarial regularization which generates virtual examples that do not exist, our adversarial examples are picked from the local entity set $V$ that are of concrete meanings. Specifically, given a KG query $(h, r, ?, m)$ and ground-truth entity $t$, \method treats entities that are textually similar to $t$ as adversarial examples. We refer these samples as \emph{Local Adversarial Regularization} (LAR) entities. To allow efficient training, we define LAR samples as the ones sharing the common tokens in entity names and descriptions with $t$, enabling us to pre-compute these LAR samples before training. This is different from previous works~\cite{FGM, PGD, FGSM} that generate virtual adversarial examples using training perturbation with large computational costs. Specifically, the LAR training objective is:
\begin{equation}
\begin{aligned}
\mathcal{L}_{l}(h, r, t, m) = \max(f(h, r, t, m)  \\
- \frac{1}{n}\sum\limits_{i=0}^{n} f(h, r, t^\Delta_{i}, m) + \gamma, 0) \label{eq:lar in} 
\end{aligned}
\end{equation}
where $t^\Delta_{i}$ is an sampled LAR entity of $t$, $\gamma$ is the margin hyperparameter, $n$ is the number of sampled LAR entities.
\input{tables/main_result_skgc}
\subsection{ Training and Inference}
\label{objective}
For training, we leverage the standard cross entropy loss with label smoothing and LAR:
\begin{equation}
\begin{aligned}
\mathcal{L}_{c}(h, &r, t, m) = - (1 - \epsilon)\cdot \log p(t|h, r, m)  \\ 
&- \frac{\epsilon}{|V|}\sum\limits_{t^{\prime} \in V/t} \cdot \log p(t^{\prime}|h, r, m)
\end{aligned}
\end{equation} 
\begin{equation}
\begin{aligned}
\mathcal{L} = \sum\limits_{(h, r, t, m) \in T}\mathcal{L}_{c}(h, r, t, m) + \alpha \cdot \mathcal{L}_{l}(h, r, t, m)
\end{aligned}
\end{equation}
where $p(t|h, r, m) = \frac{\exp{f(h, r, t, m)}}{\sum_{t^{\prime}\in V}\exp{f(h, r, t^{\prime}, m)}}$, $\epsilon$ is the label smoothing value and $\alpha$ is the LAR term weight. For inference, \method first computes the representations for KG query $(h, r, ?, m)$, then uses the entity embeddings in $G$ to compute the entity ranking. While other PLM-Based KGC models such as StAR~\cite{StAR} requires $|V|$ PLM forward pass computation for entity embeddings. Thus, \method is more computationally efficient than these baselines (See Sec.~\ref{sec:model efficiency}).

%% file: tables/main_result_skgc.tex
\begin{table*}[!t]
	\centering
	\resizebox{\textwidth}{!}{
	\begin{tabular}{lcccccccccccccccc}
		\toprule
		& \multicolumn{4}{c}{\textbf{\datawnrr{}}} &
		\multicolumn{4}{c}{\textbf{\datafb{}}} &
		\multicolumn{4}{c}{\textbf{\datawiki{}}}\\ 
		\cmidrule(r){2-5}  \cmidrule(r){6-9} \cmidrule(r){10-13} 
&MRR &H@1 &H@3 &H@10 &MRR &H@1 &H@3 &H@10 &MRR &H@1 &H@3 &H@10\\
		\midrule
\textbf{\emph{\footnotesize{Graph-Based Methods}}}\\
TransE~\cite{TransE} &.243 &.043 &.441 &.532 &.279 &.198 &.376 &.441 &.253 &.170 &.311 &.392\\
DistMult~\cite{DistMult} &.444 &.412 &.470 &.504 &.281 &.199 &.301 &.446  &.253 &.209 &.278 &.334\\
ComplEx~\cite{ComplEx} &.449 &.409 &.469 &.530 &.278 &.194 &.297 &.450 &.308 &.255 &- &.398\\
ConvE~\cite{ConvE} &.456 &.419 &.470 &.531 &.312 &.225 &.341 &.497 &- &- &- &-\\
RotatE~\cite{RotatE} &.476 &.428 &.492 &.571 &.338 &.241 &.375 &.533 &.290 &.234 &.322 &.390\\
CompGCN~\cite{CompGCN} &.479 &.443 &.494 &.546 &.355 &.264 &.390 &.535 &- &- &- &-\\
\midrule
\textbf{\emph{\footnotesize{PLM-Based Methods}}} \\ 
KG-BERT~\cite{KG-BERT} &.216 &.041 &.302 &.524 &- &- &- &.420 &- &- &- &-\\
MTL-KGC~\cite{MTL-KGC} &.331 &.203 &.383 &.597 &.267 &.172 &.298 &.458 &- &- &- &-\\
StAR~\cite{StAR} &.401 &.243 &.491 &\textbf{.709} &.296  &.205 &.322 &.482 &- &- &- &-\\
MLMLM~\cite{MLMLM} &.502 &.439 &.542 &.611 &- &- &- &- &.223 &.201 &.232 &.264 \\
KEPLER~\cite{KEPLER} &- &- &- &- &- &- &- &- &.210 &.173 &.224 &.277 \\
GenKGC~\cite{GenKGC} &- &.287 &.403 &.535 &- &.192 &.355 &.439 &- &- &- &- \\
KGT5~\cite{KGT5} &.508 &.487 & - &.544 &.276 &.210 & - &.414 &.300 &.267 &.318 &.365 \\
KG-S2S ~\cite{KGS2S} &.574 &\textbf{.531} &.595 &.661 &\underline{.336} &\underline{.257} &\underline{.373}  &.498 &- &- &- &- \\
\midrule
\method &\textbf{.575} &\underline{.522} &\textbf{.596} &\underline{.678} &\textbf{.358} &\textbf{.269} &\textbf{.393} &\textbf{.538} &\textbf{.380} &\textbf{.343} &\textbf{.399} &\textbf{.446} \\ 
\bottomrule
\addlinespace
\end{tabular}
}
\caption{Experimental results of different baseline methods on the SKGC datasets. \datawnrr and \datafb results are taken from~\citet{StAR}. \datawiki results are taken from ~\citet{KGT5}. The best PLM-based method results are in bold and the second best results are underlined.}
\label{tab:main result skgc} 
\end{table*}

%% file: sections/5.Result.tex
\section{Experiments}
In this section, we first compare \method with other competitive baselines in the SKGC and TKGC benchmarks in Sec.~\ref{sec:main result}. We then conduct ablation studies to verify the effectiveness of our propose components in \method in Sec.~\ref{sec:ablation}. We further show the efficiency and flexibility of \method in Sec.~\ref{sec:model efficiency} and~\ref{sec:flexibility}, respectively.

\paragraph{Dataset} \datawnrr~\cite{ConvE} and \datafb~\cite{FB15k237} are the most popular SKGC benchmarks where all inverse relations are removed to avoid data leakage. \datawiki~\cite{KEPLER} is a recently proposed large-scale SKGC benchmark.
For TKGC, we use \dataicews~\cite{TA-TransE} and \dataicewss~\cite{TA-TransE} which include political facts from the Integrated Crisis Early Warning System~\cite{DVN/28075_2015}. More dataset details are shown in Table~\ref{tab:dataset}.

\paragraph{Implementation Details}
All the experiments are conducted on a single GPU (RTX A6000). We tune the learning rate $\eta \in \{1$e$-3, 5$e$-4, 1$e$-4\}$, batch size $\mathcal{B} \in \{128, 256, 384, 450\}$, prompt length $\mathcal{P}_{l} \in \{2, 5, 10\}$ and LAR term weight $\alpha \in \{0.0, 0.1, 0.2\}$. While $\alpha > 0$, we employ 8 LAR samples for each training instance and gradually increase the LAR term weight from 0 to $\alpha$ using a step size of $\alpha_{step} =$ $1$e$-5$. \method uses the BERT-Large~\cite{bert} and ConvE~\cite{ConvE} model. We set the label smoothing to 0.1 and optimize \method with AdamW~\cite{AdamW}. We choose the checkpoints based on the validation mean reciprocal rank (MRR). We follow the \textit{filtered setting} in~\citet{TransE} to evaluate our model. Detailed model hyperparameters for each dataset are shown in Appendix \ref{app:implementation details}.



\input{tables/main_result_tkgc}

\subsection{Main result}
\label{sec:main result}
Table~\ref{tab:main result skgc} and Table~\ref{tab:main result tkgc} present the main SKGC and TKGC results, respectively, which demonstrate statistical significance (t-student test, $p < 0.05$). 
\paragraph{Results on SKGC} 
As for the popular medium-sized KGC benchmarks, \method achieves state-of-the-art or competitive performance compared with PLM-based KGC models. In particular, on \datafb, \method consistently outperforms all PLM-based KGC models and achieves 6.5\% (from 0.336 to 0.358) relative MRR improvement. These PLM-based baselines are all fully fine-tuned, indicating the importance of using parameter-effective prompts in the KGC task. Compared with graph-based methods, \method outperforms baseline methods by a large margin on \datawnrr(i.e. 0.575 \emph{v.s.} 0.479 on MRR)  and on \datafb(i.e. 0.358 \emph{v.s.} 0.355 on MRR). Noted that the improvement on \datafb is barely comparable to that on \datawnrr, and this discrepancy can be explained by the existence of Cartesian Product Relations (CPRs) in \datafb, which are noisy and semantically meaningless relations ~\cite{KGS2S, fb15k-237n, CVT}.  
On the \datawiki benchmark, \method significantly outperforms previous methods, showing the advantages of \method on the large-scale KGs. These results verify that with frozen PLM and accordingly much less trainable parameters, \method can achieve remarkable performance on various KGs with different scales.



\paragraph{Results of TKGC} Table~\ref{tab:main result tkgc} reports the experiment results on the \dataicews and \dataicewss benchmarks. On \dataicews, \method substantially outperforms existing TKGC methods (e.g., at least 0.03 MRR higher than previous works). On \dataicewss, \method is 0.028 and 0.045 higher than the best TKGC methods in terms of MRR and H@1, though being slightly worse on H@10 than Tero and ATiSE. On both benchmarks, \method sets new state-of-the-art performance. Note that the TKGC baseline models are often specifically designed and optimized for the TKGC task, while the only modification to \method is to add timestamp into its input. This further shows that our proposed \method method is a generally strong solution for various of KGC tasks.






\subsection{Ablation Studies}
\label{sec:ablation}
\input{sections/6.Discussion}

\subsection{Model Efficiency}
\label{sec:model efficiency}
Table~\ref{tab:modelefficiency} shows the model efficiency for \method and other PLM-based KGC methods on a single RTXA6000 GPU. \method requires much less training and evaluation time. 
\input{tables/model_efficiency}
Compared with KG-BERT~\cite{KG-BERT} and StAR~\cite{StAR}, \method is 10x faster in training and 100x faster in evaluation. This is because both KG-BERT and StAR require the PLM outputs to represent all KG entities, which introduces significant computational cost. In contrast, \method only applies BERT to represent the input queries and directly uses entity embedding matrix to compute entity ranking. We also compare \method with GenKGC~\cite{GenKGC} and KG-S2S ~\cite{KGS2S}, recently proposed PLM-based Sequence-to-Sequence KGC models. They directly generate the correct entity names and does not require to use the outputs of PLMs to represent large-scale KG entities. However, it has to maintain a huge search space for the entity names during inference and becomes much slower than \method (e.g., 0.2m \emph{vs.} 104m and 115m). In summary, \method maintains higher-level efficiency (as well as performance) compared to other PLM-based KGC methods with similar model size.


\subsection{Flexibility to Graph-based KGC models}
\label{sec:flexibility}

As we discussed in Sec.~\ref{kgcmodel}, \method is able to incorporate other graph-based KGC methods. To verify the flexibility of \method, we replace the ConvE with another two popular graph-based KGC methods: TransE and DistMult. As shown in Table~\ref{tab:model flexibility}, \method can always improve the KGC task performance after integrating with TransE, DistMult and ConvE. This indicates that \method successfully incorporate the text information into these graph-based KGC models. In particular, \method with TransE achieves a 2x improvement on MRR (from .243 to .499) and  10x improvement on H@1 (from .043 to .462). In short, \method is capable of fusing its textual knowledge with the structural knowledge provided by various of graph-based KGC models.



\input{tables/model_flexibility}


\subsection{Case Study}
In this section, we showcase how \emph{Conditional Soft Prompt} could prevent \method from over-fitting to textual information. Table~\ref{tab:case study} lists the top two entities ranked by \method and \method w/o \emph{Conditional Soft Prompt} (i.e., \method w/ FT in Table~\ref{tab:adversarial}). In the first case, \method produces two different occupations that are relevant to the \emph{whaler} in the KG Query, whilst \method w/o \emph{Conditional Soft Prompt} ranks two sea animal names as the outputs. This could be caused by the surface keywords \emph{seaman} and \emph{ship} in the KG Query. In the second case, the expected entity should be an award for the band \emph{Queen}. \method successful pick up the correct answer from many award entities using the existing KG structures, while \method w/o \emph{Conditional Soft Prompt} confuses in those candidates which are textually similar and unable to rank the ground-truth entity into top-2. In summary, \method maintains a balance between textual and structural knowledge, while \method w/o \emph{Conditional Soft Prompt} often focuses too much on the textual information in the KG Query. 

\input{tables/case_study}

%% file: tables/main_result_tkgc.tex
\begin{table*}[!htbp]
	\centering
	\begin{small}
	\resizebox{0.9\textwidth}{!}{
	\begin{tabular}{lcccccccccccc}
		\toprule
		& \multicolumn{4}{c}{\textbf{ICEWS14}} &
		\multicolumn{4}{c}{\textbf{ICEWS05-15}}  \\ 
		\cmidrule(r){2-5}  \cmidrule(r){6-9} 
&MRR &H@1 &H@3 &H@10 &MRR &H@1 &H@3 &H@10 \\
		\midrule
\textbf{\emph{\footnotesize{Graph-Based Methods}}} \\
TTransE~\cite{TTransE} &.255 &.074 &- &.601 &.271 &.084 &- &.616\\
HyTE~\cite{HyTE} &.297 &.108 &.416 &.655 &.316 &.116 &.445 &.681\\
ATiSE~\cite{ATiSE} &.550 &.436 &.629 &.750 &.519 &.378 &.606 &\underline{.794}\\
DE-SimplE~\cite{DE-SimplE} &.526 &.418 &.592 &.725 &.513 &.392 &.578 &.748\\
Tero~\cite{Tero} & .562 &.468 &.621 &.732 &.586 &.469 &\underline{.668} &\textbf{.795}\\
TComplEx~\cite{TNTComplEx} &.560 &.470 &.610 &.730 &.580 &.490 &.640 &.760\\
TNTComplEx~\cite{TNTComplEx} &.560 &.460 &.610 &.740 &\underline{.600} &\underline{.500} &.650 &.780 \\
T+TransE~\cite{TKGCframework}&.553 &.437 &.627 &\underline{.765} &- &- &- &-\\
T+SimplE~\cite{TKGCframework} &.539 &.439 &.594 &.730 &- &- &- &-\\
\midrule
\textbf{\emph{\footnotesize{PLM-Based Methods}}} \\ 
KG-S2S ~\cite{KGS2S} &\underline{.595} &\underline{.516} &\underline{.642} &.737 &- &- &- &- \\
\midrule
\method &\textbf{.628} &\textbf{.548} &\textbf{.677} &\textbf{.773} &\textbf{.628} &\textbf{.545} &\textbf{.678} &.783 \\ 
\bottomrule
\addlinespace
\end{tabular}
}
\caption{Experimental results of different baseline methods on the TKGC datasets. The results of baseline are obtained from original papers.}
\label{tab:main result tkgc} 
\end{small}
\end{table*}

%% file: sections/6.Discussion.tex
\begin{figure}[!ht] 
    \centering 
    \includegraphics[width=\columnwidth]{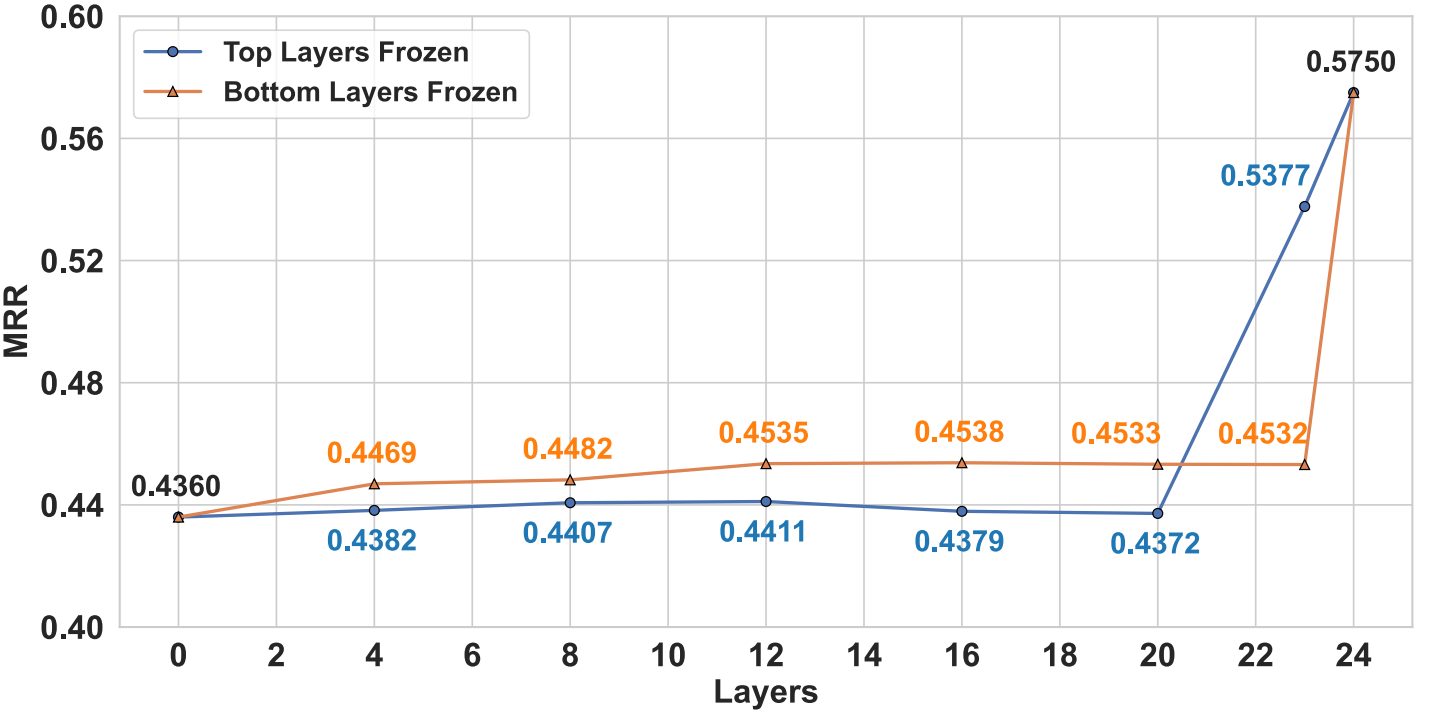}
    \caption{The effect of parameter frozen on \datawnrr. Orange and Blue lines indicate the performance when freezing parameters from bottom and top layers in PLM. The X-axis shows the number of frozen layers and the Y-axis shows the corresponding performance MRR.} 
    \label{fig:finetuning variants} 
\end{figure}

We conduct ablation study to show the effectiveness of our proposed components on \datawnrr. Table~\ref{tab:adversarial} and Figure~\ref{fig:finetuning variants} summarize the ablation study results.

\input{tables/adversarial}

\begin{figure}[!ht] 
    \centering 
    \includegraphics[width=\columnwidth]{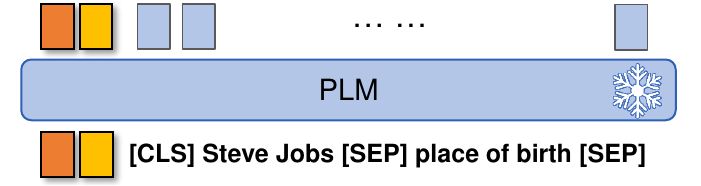}
    \caption{\emph{Joint Strategy} used in \method.} 
    \label{fig:structure variants joint} 
\end{figure}

\paragraph{KG Query Structure}
As we discussed in Sec.~\ref{methodsec}, for each KG Query $(h, r, ?, m)$, we \emph{jointly} concatenate their textual information and feed them into the frozen PLM (as shown in Figure~\ref{fig:structure variants joint}). 
\begin{figure}[!ht] 
    \centering 
    \includegraphics[width=\columnwidth]{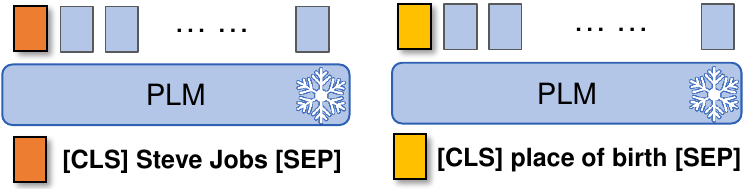}
    \caption{\emph{Separated Strategy} used in the ablation study.} 
    \label{fig:structure variants} 
\end{figure}
To demonstrate the effectiveness of this design choice, we replace it with a \emph{Separated Strategy} that is similar to the Siamese network used in~\citet{StAR}. That is, as shown in Figure~\ref{fig:structure variants}, we separately encode the textual information of $h$ and $r$ using PLMs. Table~\ref{tab:adversarial} Line 2 shows the performance of this \emph{Separated Strategy}. Compared to \method, the performance drops by 0.055 on MRR and 0.056 on H@10. The mixture of soft prompts and text representation concatenation increase the interaction between entity and relations, allowing better representation of KG Query.


\paragraph{Role of Graph-based KGC Models}
Table~\ref{tab:adversarial} Line 3 shows the performance of \method without any graph-based KGC models. For this ablation, we directly use the outputs of PLM to predict the target entity. We observe that removing this graph-based KGC model leads to a performance drop (i.e., by 0.030 MRR and 0.033 H@10). This shows that even after the complex interaction in the PLMs, an appropriate graph-based KGC model could still provide additional useful structural knowledge. This experiment verifies the necessity of combining PLM-based and graph-based KGC models together.


\paragraph{Soft Prompt Design} ~\citet{Prompt-tuning} recently propose another \emph{Soft Prompt} variant which puts longer trainable vectors at the bottom input layer. We refer it as \emph{non-layer-wise Soft Prompt}. Table~\ref{tab:adversarial} Line 4 shows the performance using this variant on \datawnrr. \method with \emph{layer-wise soft prompt} model outperforms the non-layer-wise counterpart by a large margin (i.e., 0.053 MRR and 0.066 H@10), which suggests that the layer-wised \emph{Soft Prompt} is more effective on KGC tasks. This could be explained by the fact that, to maintain similar trainable parameters, non-layer-wised \emph{Soft Prompt} requires much longer prompt vector sequences at the input, while self-attention modules are often ineffective when handling long sequences~\cite{NEURIPS2020_c8512d14}.


\paragraph{Local Adversarial Regularization} Table~\ref{tab:adversarial} Lines 5 to 8 show the ablation for adversarial regularization. 
Line 5 shows \method without LAR falls behind the full \method model by 0.041 MRR, indicating the important of LAR. From Lines 6, 7, 8, we investigate the importance of LAR entity source. We observe that \method with LAR entities that share common keywords (in name or description) outperforms the one with random LAR entities, indicating the importance of selecting appropriate  adversarial examples.


\paragraph{PLM Training Strategy}
We empirically verify the effect of freezing PLM in \method. Table~\ref{tab:adversarial} Lines 9 - 12 show the performance of \method with different level of parameter frozen. In general, the more trainable parameters in \method, the poorer \method performs. \method w/ fully fine-tuned drops significantly, by 0.138 MRR (Line 12). We further show the changes of performance as we increase the number of trainable parameters of the PLMs in Figure~\ref{fig:finetuning variants}. We freeze the PLM parameters starting from bottom layers (\textcolor{orange}{orange}) and starting from top layers (\textcolor{blue}{blue}). Both experiments suggest that the performance of \method remains nearly unchanged until the freezing operations are applied to the last few layers. As most of the layers frozen, the performance of \method grows dramatically. Interestingly, we find freezing parameters from bottom layers performs slightly better than from top layers. This could be because lower layers in BERT could capture low-level semantics (e.g., phrase features) and this information is more beneficial to the KGC task. In summary, the frozen PLM prevents \method from over-fitting the KG textual information, and therefore allows \method to achieve substantial improvements in KGC tasks.



\paragraph{Ensemble Model} 
\method has successfully combined both textual and structure knowledge for KGC using \emph{Conditional Soft Prompt}. To show the effectiveness of this design choice, we adopt a straightforward full-sized bagging strategy to combine the prediction from a graph-based KGC model and a PLM-based KGC model. We separately run the ConvE model and BERT model used in \method (i.e., same configuration for fair comparsion) and use the averaged results from both models. Table~\ref{tab:adversarial} Line 13 shows that this ensemble model is far less effective than \method. We believe this is because the ensemble model cannot deeply fuse structural and textual information like our proposed conditional soft-prompt.

\paragraph{Prompt Length}
As shown in Table~\ref{tab:prompt length study}, we conduct extensive studies to examine the impact of prompt length for \method. We observe that as the prompt length increases, there is a proportional rise in both memory and computational requirements. However, the corresponding improvement in performance is marginal. Moreover, a further increase in prompt length presents  considerable challenges in training the prompt model, leading to a decline in performance. 
\input{tables/prompt_length}

Furthermore, we conduct an investigation involving the utilization of a fully fine-tuned BERT to represent the input head entity and relation, without using prompt learning or a graph-based models. However, we find instability during the training process of this model, and consequently, the resulting model achieve very low performance compared to the results reported above.

%% file: tables/adversarial.tex
\begin{table}[!htbp]
	\centering
	\resizebox{\columnwidth}{!}{
	\begin{tabular}{clccc}
		\toprule
No. &Model &MRR &H@1 &H@10 \\
\midrule\midrule
1 &\method &.575 &.522 &.678 \\ 
2 &\method w/ \emph{Separated Strategy} &.520 &.470 &.622 \\ 
3 &\method w/o Graph KGC model &.545 &.495 &.645 \\ 
4 &\method w/ non-LW Soft Prompt &.522 &.473 &.612 \\ 
\midrule\midrule
5 & \method w/o LAR &.534 &.489 &.624 \\ 
6 &\method w/ LAR from Name &.557 &.513 &.643 \\ 
7 &\method w/ LAR from Description &.551 &.501 &.647 \\ 
8 &\method w/ Random LAR &.545 &.500 &.630 \\ 
\midrule\midrule
9 &\method w/ the last layer tunable &.537 &.494 &.621 \\ 
10 &\method w/ the last 4 layers tunable &.437 &.410 &.488 \\ 
11 &\method w/ the last 6 layers tunable &.441 &.415 &.493 \\ 
12 &\method w/ fully finetune &.436 &.409 &.484 \\ 
\midrule\midrule
13 & Ensemble model &.481 &.549 &.630  \\
\bottomrule
\addlinespace
\end{tabular}
}
\caption{Ablation Study regarding important components in \method on the benchmark of \datawnrr.} 
\label{tab:adversarial} 
\end{table}


%% file: tables/prompt_length.tex
\begin{table}[!htbp]
	\centering
	\resizebox{\columnwidth}{!}{
	\begin{tabular}{l|llll|ll}
		\toprule
        length &MRR &H@1 &H@3 &H@10 &T/EP &\#Trainable\\
		\midrule\midrule
        10 &.575 &.522 &.596 &.678 &12min &28M \\
        \midrule
        50 &.577 &.523 &.601 &.680 &23min &104M \\
        \midrule
        100 &.434 &.419 &.450 &.483 &41min &200M \\
\bottomrule
\addlinespace
\end{tabular}
}
\caption{Prompt length study of \method on \datawnrr}
\label{tab:prompt length study} 
\end{table}

%% file: tables/model_efficiency.tex
\begin{table}[!ht]
	\centering
	\resizebox{\columnwidth}{!}{
    	\begin{tabular}{l|ccrrr}
    		\toprule
    		Method  &PLM &\#Total&\#Trainable &T/Ep &Inf\\
    		\midrule
    		\midrule
            \multirow{2}{*}{KG-BERT}  &RoBERTa base  &125M &125M &79m &954m\\
             &RoBERTa large &355M &355M &142m &2928m\\
            \midrule
            \multirow{2}{*}{StAR}  &RoBERTa base &125M &125M &42m &27m\\
              &RoBERTa large &355M &355M &103m &34m\\
            \midrule
            \multirow{2}{*}{GenKGC}  &BART base &140M &140M &5m &88m\\
             &BART large &400M &400M &11m &104m \\
            \midrule
            \multirow{2}{*}{KG-S2S} &T5 base& 222M& 222M& 10m& 81m\\
            &T5 large& 737M& 737M& 27m& 115m \\
            \midrule
            \midrule
            \multirow{2}{*}{\method} &BERT base &126M &17M &4m &0.1m \\
             &BERT large &363M &28M &12m &0.2m \\
            \bottomrule
            \addlinespace
        \end{tabular}
    }
    \caption{Comparisons of model efficiency for \method and other PLM-based methods on \datawnrr with FP32 precision. \#Total and \#Trainable denotes the total and trainable parameters, respectively. T/Ep and Inf denotes the training time per epoch and inference time.}
    \label{tab:modelefficiency}
\end{table}

%% file: tables/model_flexibility.tex
\begin{table}[!htbp]
	\centering
	\resizebox{\columnwidth}{!}{
	\begin{tabular}{lllll}
		\toprule
        Methods &MRR &H@1 &H@3 &H@10 \\
		\midrule\midrule
        TransE&.243 &.043 &.441 &.532 \\
        \quad + \method &$.499_{\uparrow .256}$ &$.462_{\uparrow .419}$ &$.515_{\uparrow .074}$ &$.569_{\uparrow .037}$ \\ 
        \midrule
        DistMult&.444 &.412 &.470 &.504 \\
        \quad + \method&$.543_{\uparrow .099}$ &$.494_{\uparrow .082}$ &$.562_{\uparrow .092}$ &$.639_{\uparrow .135}$ \\ 
        \midrule
        ConvE &.456 &.419 &.470 &.531 \\
        \quad + \method&$.575_{\uparrow .119}$ &$.522_{\uparrow .103}$ &$.596_{\uparrow .126}$ &$.678_{\uparrow .147}$  \\ 
\bottomrule
\addlinespace
\end{tabular}
}
\caption{\datawnrr results of \method with different graph-based methods.}
\label{tab:model flexibility} 
\end{table}

%% file: tables/case_study.tex
\begin{table}[!htbp]
	\centering
	\resizebox{\columnwidth}{!}{
	\begin{tabular}{l}
		\hline
\rowcolor{mygray}
\textbf{KG Query}: \\
\rowcolor{mygray}
\quad whaler [a seaman who works on a ship that hunts whales] | hypernym \\
\hline
\method:\\
\quad $A1^{*}$: tar [a man who serves as a sailor] \\
\quad A2: crewman [a member of a flight crew] \\
\midrule
\method w/o \emph{Conditional Soft Prompt}:\\
\quad A1: pelagic bird [bird of the open seas] \\
\quad A2: mackerel [any of various fishes of the family scombridae] \\
\hline
\rowcolor{mygray}
\textbf{KG Query}: \\
\rowcolor{mygray}
\quad Queen [queen are a british rock band formed in london in 1970 ...] | award \\
\hline
\method:\\
\quad $A1^{*}$: Grammy Award for Best Pop Performance by Group with Vocal [\textit{...}] \\
\quad $A2$: MTV Video Music Award for Best Visual Effects [\textit{the following is  ...}] \\
\midrule
\method w/o \emph{Conditional Soft Prompt}:\\
\quad $A1$: Grammy Award for Best Music Film [\textit{the grammy award for best  ...}] \\
\quad $A2$: Razzie Award for Worst Original Song [\textit{the razzie award for worst...}]\\

\bottomrule
\addlinespace
\end{tabular}
}
\caption{Case study of \method. Texts in brackets are entity descriptions. $*$ denotes ground-truth entity.} 
\label{tab:case study} 
\end{table}

%% file: sections/7.Conclusion.tex
\section{Conclusion and Future Work}
In this paper, we propose \method, a PLM-based KGC model that 
effectively fuses the KG structural knowledge and avoids over-fitting towards textual information. The key innovation of \method is the \emph{Conditional Soft Prompt} that connects between a graph-based KGC models and a frozen PLM avoiding the textual over-fitting issue. We conduct experiments on five popular KGC benchmarks in SKGC and TKGC settings and the results show that \method outperforms several strong graph-based and PLM-based KGC models. We also show the efficiency and flexibility of \method. For future work, we plan to adapt our method to other relevant knowledge-intensive downstream tasks, such as fact checking and open-ended question answering.

%% file: sections/9.Limitations.tex
\section{Limitations}
\method successfully integrates both graph-based and textual representations in the KGC task, achieving substantial performance and efficiency improvement. However, similar to other PLM-based methods, this comes at the cost of increased computational resources (v.s. graph-based KGC models). In addition, we find that \method may occasionally collapse on small KGC benchmarks (e.g. \datawnrr{}) under specific random seeds. This is probably due to the nature of \emph{Soft Prompts}, which involve much smaller number of trainable parameters, compared to fine-tuned models. However, we never see similar phenomena when training \method in the large KGC benchmarks (e.g., \datawiki{}). We plan to solve these issues for \method as future work.



%% file: sections/10.Acknowledgement.tex
\section*{Acknowledgement}
We thank the anonymous reviewers for their insightful suggestions to improve this paper. This research / project is supported by the National Research Foundation, Singapore and Infocomm Media Development Authority under its Trust Tech Funding Initiative and A*STAR SERC Central Research Fund (UIBR). Any opinions, findings and conclusions or recommendations expressed in this material are those of the author(s) and do not reflect the views of National Research Foundation, Singapore and Infocomm Media Development Authority. 

%% file: sections/8.Appendix.tex
\section{Dataset}
We use SKGC datasets released from ~\cite{KG-BERT} and TKGC datasets from ~\cite{TA-TransE}. We follow the original split in our experiments. Table \ref{tab:dataset} shows the statistics of the datasets. All of these datasets are open-source English-written sources without any offensive content. They are introduced only for research use.
\input{tables/dataset}

\section{Hyperparameters} \label{app:implementation details}
Hyperparameters are selected with grid search on the validation set. The optimal hyperparameters are presented in Table~\ref{tab:hyperparameters}
\input{tables/hyperparameters}

\section{Baseline Methods}
\method is compared against a variety of state-of-the-art baseline methods on SKGC and TKGC tasks. For SKGC, we include popular graph-based methods, i.e. TransE~\cite{TransE}, DistMult~\cite{DistMult}, ComplEx~\cite{ComplEx}, ConvE~\cite{ConvE}, RotatE~\cite{RotatE} and CompGCN~\cite{CompGCN}. We also compare \method against several competitive PLM-based methods, i.e. KG-BERT~\cite{KG-BERT}, MTL-KGC~\cite{MTL-KGC}, StAR~\cite{StAR}, MLMLM~\cite{MLMLM}, KEPLER~\cite{KEPLER}, GenKGC~\cite{GenKGC}, KGT5~\cite{KGT5} and KG-S2S~\cite{KGS2S}. For TKGC, we compare \method with graph-based TKGC baselines, including: TTransE~\cite{TTransE}, HyTE~\cite{HyTE}, ATiSE~\cite{ATiSE}, DE-SimplE~\cite{DE-SimplE}, Tero~\cite{Tero}, TComplEx~\cite{TNTComplEx}, TNTComplEx~\cite{TNTComplEx}, T+TransE~\cite{TKGCframework}, T+SimplE~\cite{TKGCframework}. PLM-based baselines for TKGC includes KG-S2S~\cite{KGS2S}

%% file: tables/dataset.tex
\begin{table}[!htbp]
	\centering
	\resizebox{\linewidth}{!}{
	\begin{tabular}{lccccc}
\toprule
Dataset  &$|\mathcal{E}|$ & $|\mathcal{R}|$ & |Train| & |Valid| &|Test|  \\
\midrule
\textbf{\emph{\footnotesize{SKGC}}}\\
\datawnrr &40,943 &11 &86,835 &3,034 &3,134 \\
\datafb &14,541 &237 &272,115 &17,535 &20,466 \\
\datawiki & 4,594,485 &822 &20,614,279 & 5,163 & 5,133\\
\midrule
\textbf{\emph{\footnotesize{TKGC}}}\\
\dataicews &6,869 &230 &72,826 &8,941 &8,963 \\
\dataicewss &68,544 &358 &189,635 &1,004 &2,158 \\
\bottomrule

\end{tabular}
}
\caption{Statistics of the Datasets.}
\label{tab:dataset} 
\end{table}

%% file: tables/hyperparameters.tex
\begin{table}[!htbp]
	\centering
	\resizebox{0.8\linewidth}{!}{
	\begin{tabular}{lcccc}
\toprule
Dataset  &$\eta$ &$\mathcal{B}$ &$\mathcal{P}_{l}$ &$\alpha$ \\
\midrule
\datawnrr &5$e$-4 &128 &10 &0.1\\
\datafb &5$e$-4 &128 &10 &0.1\\
\datawiki &1$e$-4 &450 &5 &0.0\\
\dataicews &5$e$-4 &384 &5 &0.1\\
\dataicewss &5$e$-4 &384 &5 &0.0\\
\bottomrule

\end{tabular}
}
\caption{Optimal hyperparameters.}
\label{tab:hyperparameters} 
\end{table}